\documentclass[lettersize,journal]{IEEEtran}
\usepackage{amsmath,amsfonts}
\usepackage{algorithmic}
\usepackage{algorithm}
\usepackage{array}
\usepackage[caption=false,font=normalsize,labelfont=sf,textfont=sf]{subfig}
\usepackage{textcomp}
\usepackage{stfloats}
\usepackage{url}
\usepackage{verbatim}
\usepackage{graphicx}
\usepackage{cite}
\usepackage{tabularx}
\usepackage{multirow}
\usepackage{colortbl}
\usepackage{bbding}
\usepackage{color}
\usepackage{adjustbox}
\usepackage{booktabs}
\usepackage{subcaption}
\usepackage{hyperref}
\hypersetup{
    colorlinks=true,
    linkcolor=blue,
    citecolor=blue, 
    urlcolor=blue  
}

\begin{document}

\title{TDS-CLIP: Temporal Difference Side Network for Efficient Video Action Recognition}

\author{Bin Wang, Wentong Li, Wenqian Wang, Mingliang Gao, Runmin Cong and Wei Zhang

\thanks{Bin Wang is with the School of Electrical and Electronic Engineering, Shandong University of Technology, Zibo, 255000, China, also with the School of Control Scienceand Engineering, Shandong University, Jinan 250061, China, also with the KeyLaboratory of Machine Intelligence and System Control, Ministry of Education,Jinan 250061, China (e-mail: dqwangbin@sdut.edu.cn).

Wentong Li is with School of Artificial Intelligence, Nanjing University of Aeronautics and Astronautics, Nanjing 210000, China (e-mail: wentong\_li@nuaa.edu.cn).

Wenqian Wang is with the Pillar of Information Systems Technology and Design, Singapore University of Technology and Design, Singapore (e-mail: wenqian\_wang@sutd.edu.sg).

Mingliang Gao is with the School of Electrical and Electronic Engineering, Shandong University of Technology, Zibo, 255000, China (e-mail: mlgao@sdut.edu.cn).

Runmin Cong and Wei Zhang are with the School of Control Scienceand Engineering, Shandong University, Jinan 250061, China, also with the KeyLaboratory of Machine Intelligence and System Control, Ministry of Education,Jinan 250061, China (e-mail: rmcong@sdu.edu.cn; davidzhang@sdu.edu.cn).

(\emph{Corresponding author: Wei Zhang.})  \\
}}



\maketitle

\begin{abstract}
Recently, large-scale pre-trained vision-language models (e.g., CLIP), have garnered significant attention thanks to their powerful representative capabilities. 
This inspires researchers in transferring the knowledge from these large pre-trained models to other task-specific models, e.g., Video Action Recognition (VAR) models, via particularly leveraging side networks to enhance the efficiency of parameter-efficient fine-tuning (PEFT). 
However, current transferring approaches in VAR tend to directly transfer the frozen knowledge from large pre-trained models to action recognition networks with minimal cost, instead of exploiting the temporal modeling capabilities of the action recognition models themselves.
Therefore, in this paper, we propose a novel memory-efficient Temporal Difference Side Network (TDS-CLIP) to balance knowledge transferring and temporal modeling, avoiding backpropagation in frozen parameter models. Specifically, we introduce a Temporal Difference Adapter (TD-Adapter), which can effectively capture local temporal differences in motion features to strengthen the model's global temporal modeling capabilities. Furthermore, we designed a Side Motion Enhancement Adapter (SME-Adapter) to guide the proposed side network in efficiently learning the rich motion information in videos, thereby improving the side network's ability to capture and learn motion information.
Extensive experiments are conducted on three benchmark datasets, including Something-Something V1\&V2, and Kinetics-400. Experimental results show that our method achieves competitive performance in video action recognition tasks. Code will be available at \href{https://github.com/BBYL9413/TDS-CLIP/}{https://github.com/BBYL9413/TDS-CLIP.}
\end{abstract}

\begin{IEEEkeywords}
Action recognition, side network, temporal modeling, parameter-efficient fine-tuning, motion information.
\end{IEEEkeywords}    
\section{Introduction}
\label{sec:intro}

\IEEEPARstart{I}{n} recent years, Large Language Models (LLMs) have achieved significant success~\cite{mann2020language,touvron2023llama,ryoo2021tokenlearner, zhou2023twinformer, wasim2023vita, wang2024vilt, shu2022multi} in the fields such as text-video retrieval~\cite{wu2023cap4video}, text-to-image generation~\cite{qu2023layoutllm}, video re-identification~\cite{liu2024video}, etc., gradually extending their influence to the domain of computer vision. Video Action Recognition (VAR) tasks, a challenging research area within computer vision, have seen mainstream large language models represented by the CLIP paradigm~\cite{clip}, which includes a Vision Transformer (ViT) vision encoder~\cite{dosovitskiy2020image} for image processing and a text encoder. CLIP establishes a strong association between images and text and is applied to various video tasks. This raises a thought-provoking question: how can we effectively utilize the pre-training of such large models for VAR tasks?

\begin{figure}[t]
    \centering
    \includegraphics[width=\linewidth]{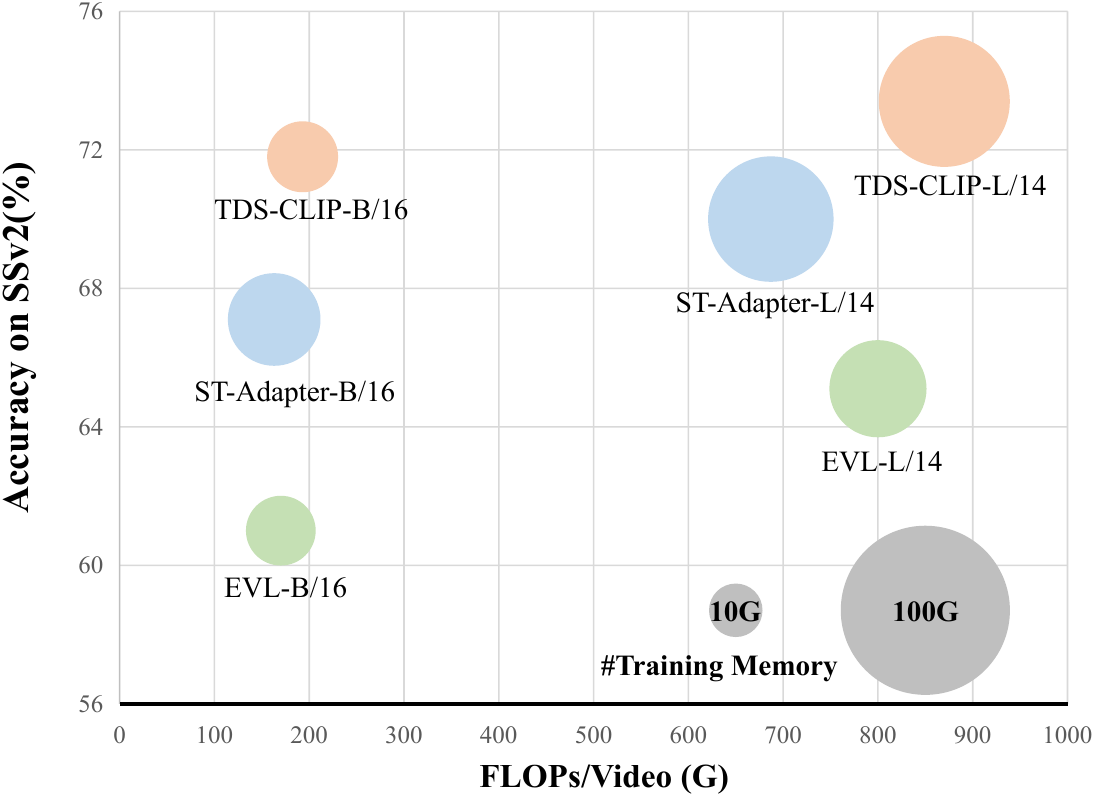}
    \caption{Comparison of per-video GFLOPs and accuracy on SSv2~\cite{goyal2017something}. The size of the bubbles represents the memory usage during training. EVL: Efficient Video Learning~\cite{lin2022frozen}. ST-Adapter: Spatio-Temporal Adapter~\cite{pan2022st}.}
    \vspace{-0.3cm}
\end{figure}

\begin{figure}[t]
    \centering
    \includegraphics[width=0.9\linewidth]{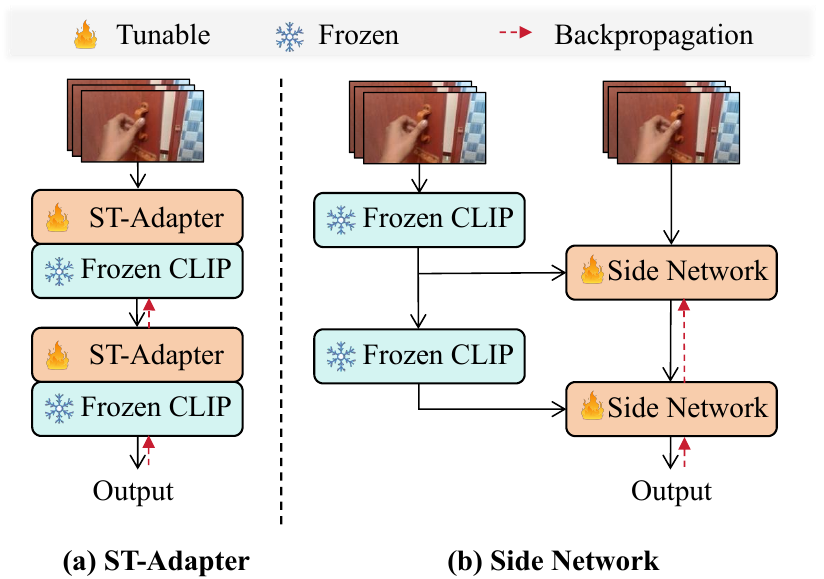}
    \caption{Comparison with commonly used and effective fine-tuning methods for video recognition. (a): ST-Adapter: Spatio-Temporal Adapter \protect\cite{pan2022st}. (b): Ladder Side-tuning \protect\cite{sung2022lst}.}
    \label{fig:stadapter}
\end{figure}

Recently, the emergence of Parameter-Efficient Fine-Tuning (PEFT) methods~\cite{houlsby2019parameter,lester2021power} has effectively alleviated this issue. PEFT adapts specific tasks by adjusting the learnability of certain parameters or introducing various adapters, avoiding extensive parameter adjustments across the entire model. With the advent of PEFT, researchers are exploring the possibility of freezing the original CLIP parameters and incorporating various adapters~\cite{wang2024multimodal} or prompts~\cite{wasim2023vita}. These methods achieve impressive results in video understanding and are widely used in tasks such as video action recognition. As shown in Fig. \ref{fig:stadapter}(a), a popular design for efficient transfer learning is to insert tunable structures between pre-trained and parameter-frozen CLIP visual transformer blocks. Although introducing adapters under frozen parameters facilitates temporal action modeling, backpropagation still passes through the frozen layers, consuming memory. So, is there a way to solve this problem effectively?

Inspired by these works~\cite{sung2022lst,qing2023disentangling,yao2023side4video, zhou2022lgnet}, we argue that independent side-network models can better uncover temporal cues without compromising pre-training.
Traditional convolutional neural networks go through a similar process of migrating pre-training from Image-Net into various variants, so we believe that CLIP can do the same. For VAR tasks, there is often a need to emphasize the effective handling of temporal branches. Previous approaches ignore the importance of independent temporal branches, and thus the potential of these approaches for temporal modeling using side networks in video understanding tasks has yet to be fully explored and exploited.

In this work, we propose a method based on CLIP fine-tuning that balances memory efficiency and temporal modeling performance, called TDS-CLIP. The idea of this work is that since there are not enough computing resources to explore the entire spatial, temporal, and fusion mechanism, we rely on CLIP's powerful spatial modeling capabilities to focus on serving temporal branches, and then combine independent temporal branches with the effective spatiotemporal modeling mechanism in traditional convolutional neural networks to form our TDS-CLIP. First, we shift the adapter positions used for the CLIP vision branch to the side network to achieve memory-efficient pre-training fine-tuning. To better model the temporal information of videos, we design a new Temporal Difference Adapter (TD-Adapter) that effectively captures local temporal differences to enhance the model's global temporal modeling capability. Second, we integrate the TD-Adapter with a multi-head self-attention module and feed-forward network to form a learnable side network module, used for parameter training and backpropagation. Finally, to improve the side network's capability of capturing video motion information, we introduce a Side Motion Enhancement Adapter (SME-Adapter) that acts on the input of the side network to guide it in effectively learning additional motion information relevant to action labels.

\begin{figure*}[t]
    \centering
    \includegraphics[width=\linewidth]{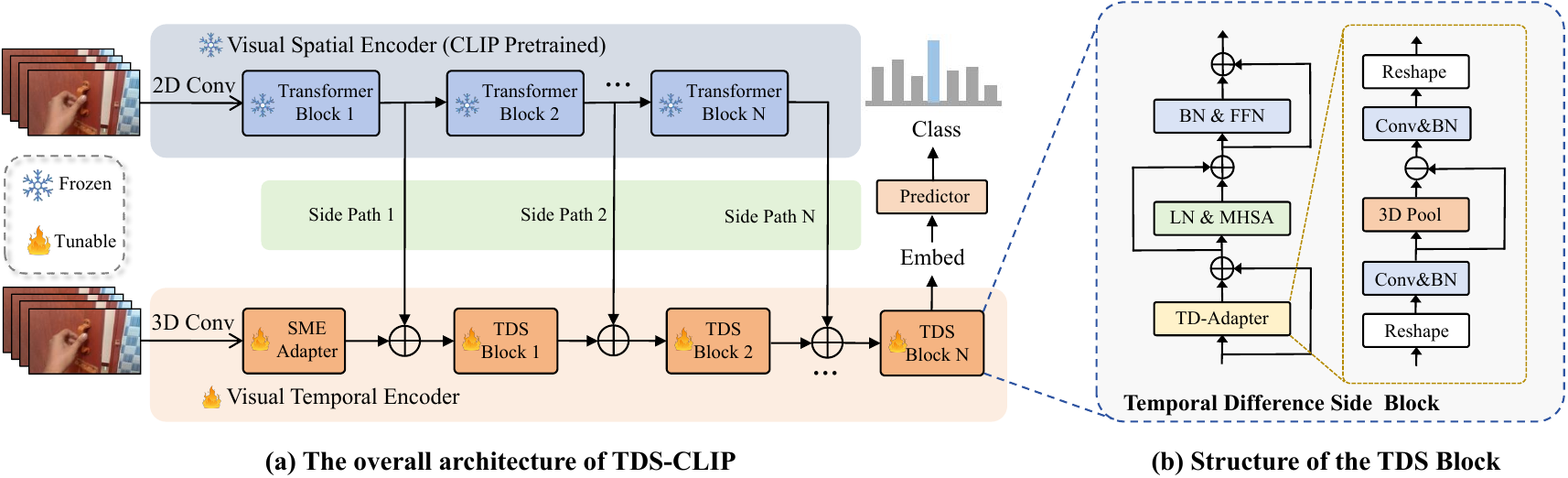}
    \caption{(a) Illustration of the proposed TDS-CLIP for video understanding. TDS-CLIP consists of a frozen CLIP visual spatial encoder and a trainable visual-temporal encoder, where the visual temporal encoder is composed of the proposed SME-Adapter and TDS Block to form the side network, and the TDS Block is composed of the proposed TD-Adapter and ViT module. (b) Detailed Structure of
    the proposed TDS Block.}
    \label{fig:overall}    
    \vspace{-0.3cm}
\end{figure*}
We evaluate the performance of our method on several large-scale temporal action recognition datasets, including Something-Something V1~\cite{goyal2017something}, Something-Something V2~\cite{goyal2017something}, and Kinetics-400~\cite{kay2017kinetics}. Experimental results demonstrate that the proposed method effectively accomplishes video action understanding tasks. 

Our contributions are summarized as follows:

\begin{itemize}
\item We propose a novel memory-efficient and temporal-branching-focused Temporal Difference Side Network for video action recognition that balances knowledge transfer and temporal modeling without requiring backpropagation in models with frozen parameters.
\item We propose the Temporal Difference Adapter (TD-Adapter) and the Side Motion Enhancement Adapter (SME-Adapter), which enhance the model's ability to capture and learn from motion information, thereby improving both local and global temporal analysis.
\item Extensive experiments evaluate the efficacy of TDS-CLIP on public datasets, demonstrating competitive performance against state-of-the-art approaches.
\end{itemize}
\section{Related Work}
\label{sec:formatting}

\textbf{VLM for Video Action Recognition.} ViT~\cite{dosovitskiy2020image} models are gradually replacing convolutional neural networks in the mainstream due to their powerful learning capabilities and flexibility~\cite{xu2021videoclip, wu2023revisiting, yu2024tf, jia2022visual, zhang2021tip, jiang2020can, herzig2022object}. EVA-CLIP~\cite{sun2023eva} proposes a 64-layer ViT-E with larger scale parameters, and the effective adaptation of such a huge image model to the video domain is a task well worth exploring. Meanwhile, Transferring pre-trained models of language-images from the ViT architecture to video attracts widespread attention~\cite{cheng2021improving}. Previous action recognition methods rely on pre-trained models from ImageNet~\cite{simonyan2014two}. Currently, powerful text-image pre-trained models based on CLIP are gradually emerging and are increasingly used in VAR tasks~\cite{wang2023actionclip}. EVL~\cite{lin2022frozen} uses parallel Transformer decoders to extract spatial features from the frozen CLIP model for VAR tasks. ST-Adapter~\cite{pan2022st} introduces a spatiotemporal convolution adapter to endow the CLIP model with the capability of understanding videos. AIM~\cite{yang2023aim} introduces spatial adaptation, temporal adaptation, and joint adaptation, gradually equipping image models with spatiotemporal reasoning abilities, further enriching the types of adapters to suit different video action recognition tasks. GPT4Ego~\cite{dai2024gpt4ego} proposes a VLM-based zero-shot egocentric action recognition method capable of exploiting the rich semantic and contextual details in self-centered videos for fine-grained conceptual and descriptive alignment, as well as designing a new selforiented visual parsing strategy thereby learning the visual contextual semantics associated with actions. DUALPATH~\cite{park2023dual} achieves dynamic modeling using spatial and temporal adaptation paths, merging successive frames into a grid-like frameset to accurately infer relationships between markers. M2-CLIP~\cite{wang2024multimodal} introduces a TED-Adapter, enhancing the model's generalization ability while maintaining video action recognition performance, and successfully uses it for multimodal VAR tasks. These methods effectively transfer the powerful pre-trained CLIP visual encoder to VAR tasks but do not address the issue of memory consumption during backpropagation.\\

\noindent\textbf{Side Network.} The side-tuning network~\cite{zhang2020side} is first proposed to address overfitting and the problem of forgetting in incremental learning, and it applies to various downstream tasks. In vision tasks, LST~\cite{sung2022lst} tackles this problem by implementing a side branch network attached to the CLIP visual model. This scheme is effective in avoiding the passing of gradients in the visual encoder of frozen parameters thus reducing memory usage. Inspired by the LST, DiST~\cite{qing2023disentangling} uses a dual encoder structure, employing the pre-trained CLIP base model as the spatial encoder and introducing a lightweight additional network as the temporal encoder, finally integrating spatiotemporal information through an ensemble branch to achieve memory-efficient video understanding. Side4Video~\cite{yao2023side4video} further explores the effective forms of memory-efficient side networks in video action recognition tasks and proposes a spatiotemporal side network for video understanding tasks. Although these methods demonstrate that using side networks can achieve memory-efficient image-to-video transfer learning, they do not deeply explore the temporal modeling capabilities of side network models, which we further enhance. However, we can see that these methods do not favor temporal branching and treat everything equally, thus limiting performance. We would like to liberate temporal branching completely and keep it fully expandable. Inspired by the powerful performance of convolutional neural networks~\cite{li2020tea, wang2021tdn, feichtenhofer2019slowfast} in temporal modeling, in this work, we introduce corresponding temporal modeling adapters at both the input and the path of the side network composed of ViT encoders, transforming it into a memory-efficient and powerful temporal modeling framework for VAR tasks. Its extensibility in temporal modeling is well preserved and its ability to reduce memory overhead by exploiting the characteristics of the measurement network provides some meaningful assistance to subsequent researchers.

\section{Method}

As shown in Fig. \ref{fig:overall}, the overall architecture of TDS-CLIP framework consists of two key components: the visual spatial encoder and the visual temporal encoder. The visual spatial encoder is composed of a frozen CLIP pre-trained ViT, which provides strong spatial semantic features for the input video frames. The visual temporal encoder is a side network structure that accepts the output features from the visual spatial encoder. It takes the dense video frame features provided by the SME-Adapter as input and transforms the spatial features from the frozen pre-trained CLIP branch into an effective form suitable for temporal modeling in video understanding.

\subsection{Denotations}
Here, we present an overview of the entire architecture and present the details of the visual spatial encoder and the visual temporal encoder in the following two sections. The input video frames to the model are formally defined as $V \in \mathbb{R}^{3\times T\times H\times W}$, where $H$ and $W$ represent spatial dimensions, and $T$ represents the sampled frames.


The spatial encoder consists of the ViT with frozen parameters from the CLIP visual branch, which can independently capture the spatial semantic features of sparse input video frames, thereby leveraging its powerful pre-training to enhance training performance. Following the ViT format, the input frames are first divided into $N=\frac{HW}{P^{2} }$  patches, each with a shape of $P^{2}$, and then projected through a fully connected layer. Then, for the input full-length video frame sequence, each patch feature vector is represented as $X_{i,j} $, where $i=\{1,2,...,T\}$ and $j=\{1,2,...,N\}$ denotes the patch index. Additionally, a learnable $E_{cls}$ token and a trainable position embedding $E_{pos}$ are added to each frame. Mathematically, the frame-level input is constructed as:
\begin{equation}
    [e^{(0)},x^{(0)}]=([E_{cls};{X_{i,j} ]+E_{pos}} ),
    \label{eq:1}
\end{equation}
with a total of $L$ transformer blocks, if we place the visual encoder before each transformer layer, the inputs will be processed sequentially, this format can be represented as:

\begin{equation}
    Z^{l}=Transformer^{l-1}([e^{(l-1)},x^{(l-1)}] ),
    \label{eq:1}
\end{equation}
where $l=\{1,2,...,L\}$ denotes the number of layers of the ViT, The output shape is $Z^{l} \in \mathbb{R}^{T\times (N+1)\times C}$.


\subsection{Visual Temporal Encoder}
The spatial semantic features from the frozen CLIP ViT can guide the model in learning the appearance and background information of video frames, but for video understanding tasks, the model needs to have temporal modeling capabilities. The visual temporal encoder is designed as a trainable side network structure that accepts the output from the visual spatial encoder, featuring its own independent video feature input mode and trainable ViT modules. Specifically, for the video frames input, we aim to guide the side network to effectively learn the latent motion information in the video through a Side Motion Enhancement Adapter (SME-Adapter). For the trainable ViT module, we design a Temporal Difference Adapter (TD-Adapter) to integrate the CLIP spatial semantic information with the temporal information from the side network, thereby assisting the model in performing effective temporal modeling.

\begin{figure}[t]
    \centering
    \includegraphics[width=0.8\linewidth]{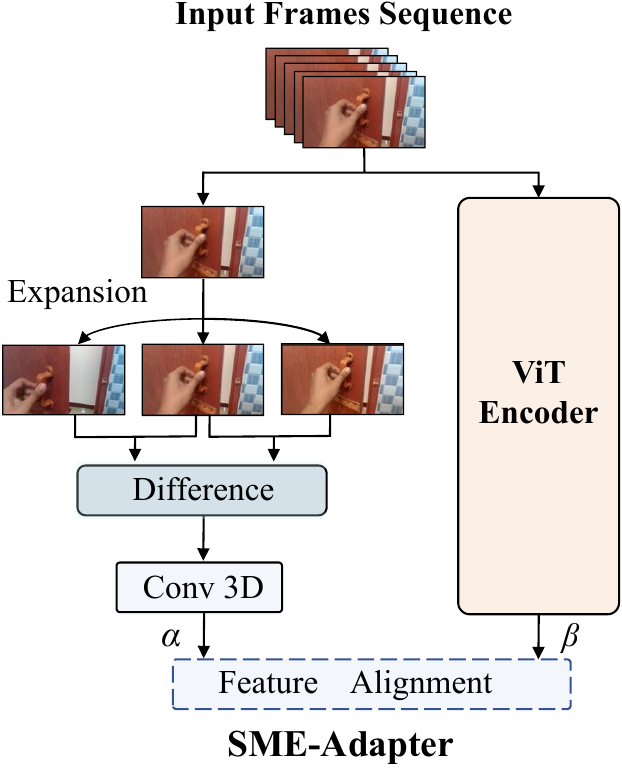}
    \caption{Detailed structure of the proposed SME-Adapter. We illustrate the feature map as the original RGB frame for better intuitive understanding. “Diff” indicates difference operation.}
    \label{fig:sme}   
  
\end{figure}
\subsubsection{SME-Adapter} Specifically, our SME-Adapter performs low-level feature extraction at the beginning of the side network, utilizing temporal differences between local RGB frames to model local motion features. As shown in Fig. ~\ref{fig:sme}, for the input video sequence $V \in \mathbb{R}^{3\times T\times H\times W}$, assuming that the i-frame is $V_i$, we first extend the sampling by $n$ frames centered on $V_i$, formally denoted as $V_{i}^n=\{  V_{i-n},...,V_{i},...,V_{i+n} \}$. Then, we extract the feature differences $D_i$ between frames within a local sampling window centered on $V_{i}^n$, and finally concatenate them along the channel dimension. This effective form is proposed as:
\begin{equation}
    D_i=Concat(V_{i-n+1} - V_{i-n},...,V_{i+n} - V_{i+n-1}),
\end{equation}
subsequently, the SME-Adapter output feature $S_{out}^{i}$ is fused into the original ViT structure in the form of residuals, while introducing factor control weights.
$S_{out}^{i}$ will be used as an input to the side network and integrated with the ViT output of the first frozen CLIP into the TDS block.
Formally, this phase can be represented as:
\begin{equation}
    S_{out}^{i} =\alpha Reshape(Conv(D_i))+\beta \mathrm{ViT} (V_i),
\end{equation}
where, the $\alpha$ and $\beta$ represent adjustment coefficients and the output after convolution needs to be reshaped into the form of ViT.

\subsubsection{TD-Adapter} To transfer the learning capability of the CLIP visual branch to video tasks, an additional temporal modeling module is required. Thus, we design a TD-Adapter in the side network, which accepts both spatial features from the frozen ViT and temporal input features from the side network, transforming them into encoder features suitable for video temporal modeling.

As shown in Fig.~\ref{fig:overall}(b), the input feature of the TD-Adapter is denoted as $Z_{v} \in \mathbb{R}^{T\times N\times C}$. First, it is reshaped to a form $Z_{s} \in \mathbb{R}^{C\times T\times H\times W}$ suitable for 3D convolution. Then, a set of convolution operations is used for dimensionality reduction to obtain the feature $Z_{s} \in \mathbb{R}^{C/r\times T\times H\times W}$, $r$ denotes rate of reduction. Next, we capture the sensitive feature values of the local temporal dimension features using a 3D max-pooling operation, and a feature difference operation is employed to obtain dynamic features of local motion patterns. Compared to conventional displacement~\cite{li2020tea}, pooling is a simpler and more effective operation.
\begin{equation}
    D_z=Z_s-Pool(Z_s),
\end{equation}
where $D_z$ denotes the feature difference. This method is commonly used in the early stages of convolutional models for efficient temporal modeling~\cite{wang2021action,li2020tea,wang2021tdn}. 

Finally, the reshaping operation and a set of convolution operations are again used to recover the feature shapes suitable for the ViT encoder.

\subsection{TDS-CLIP for video action recognition} We integrate the proposed SME-Adapter and TD-Adapter into Side4Video~\cite{yao2023side4video} architecture, where the TD-Adapter, combined with a trainable ViT encoder, forms the TDS-CLIP block. We retain the CLS token shift~~\cite{liu2022ts2,zhang2021token} operation to further improve the capacity for learning temporal information, offering the advantage of minimal additional memory overhead. The number of these blocks matches the frozen ViT encoders in the visual branch of CLIP and combined via side-connections. For action recognition tasks, we obtain the prediction scores by performing global average pooling on the output of the final TDS-CLIP block.

\begin{table*}[ht]
\label{tab:1}

\caption{\centering Comparison of performance with state-of-the-art methods on SSv1 \& SSv2. “Cr.” and “Cl.” are the abbreviation for “spatial crops” and “temporal clips”.}
\centering

\begin{tabular}{l c c c c c c c}
\toprule
\multirow{2}{*}{\textbf{Method}}  & \multirow{2}{*}{\textbf{Pre-train}} & \multirow{2}{*}{\textbf{Frames$\times$Cr.$\times$Cl.}}  & \multirow{2}{*}{\textbf{TFLOPs}} &\multicolumn{2}{c}{\textbf{SSv1(\%)}}& \multicolumn{2}{c}{\textbf{SSv2(\%)}}\\

& & & & \textbf{Top-1} & \textbf{Top-5} & \textbf{Top-1} & \textbf{Top-5}\\
\midrule
\multicolumn{8}{l}{\textit{Full Fine-tuning}} \\
ViViT-L~\cite{arnab2021vivit} & IN-21K &32$\times$3$\times$4&1.0$\times$12&-&-&65.9&89.9\\
Video Swim-B~\cite{liu2022video}& IN-21K &32$\times$3$\times$1&0.3$\times$3&-&-&69.6&92.7\\
MViT-B~\cite{fan2021multiscale}& Kinetics-600 &32$\times$3$\times$1&0.2$\times$3&-&-&68.7&91.5\\
UniFormer V2-L/14~\cite{li2023uniformerv2}& CLIP-400M &16$\times$3$\times$1&0.9$\times$3&60.5&86.5&\textbf{72.1}&\textbf{93.6}\\
ILA-ViT-L/14~\cite{tu2023implicit}& CLIP-400M &16$\times$3$\times$4&0.9$\times$12&-&-&67.8&90.5\\
S-ViT-B/16~\cite{zhao2023streaming} &CLIP-400M& 32$\times$3$\times$2& 0.3$\times$6 & - & - & 69.3 & 92.1\\
STAN-B/16~\cite{liu2023revisiting} &CLIP-400M& 16$\times$3$\times$1& 0.7$\times$1 & - & - & 69.5 & 92.7\\
\midrule
\multicolumn{8}{l}{\textit{Frozen CLIP}} \\
EVL-L/14~\cite{lin2022frozen} &CLIP-400M& 8$\times$3$\times$1& 0.8$\times$3 & - & - & 65.1 & -\\
ST-Adapter-L/14~\cite{pan2022st} &CLIP-400M& 16$\times$3$\times$1& 1.4$\times$3 & - & - & 71.9 & \textbf{93.4}\\
DUALPATH-L/14~\cite{park2023dual} &CLIP-400M& 16$\times$3$\times$1& 0.5$\times$3 & - & - & 70.2 & 92.7\\
AIM-L/14~\cite{yang2023aim} &CLIP-400M& 16$\times$3$\times$1& 1.9$\times$3 & - & - & 69.4 & 92.3\\
DiST-L/14~\cite{qing2023disentangling} &CLIP-400M& 16$\times$3$\times$1& 1.4$\times$3 & - & - & \textbf{72.5} & 93.0\\
M\textsuperscript{2}-CLIP-B/16~\cite{wang2024multimodal} &CLIP-400M& 32$\times$3$\times$1& 0.8$\times$3 & - & - & 69.1 & -\\
Side4Video-B/16~\cite{yao2023side4video} &CLIP-400M& 16$\times$3$\times$2& 0.4$\times$6 & 60.7 & 86.0 & 71.5 & 92.8\\
\midrule
\multicolumn{8}{l}{\textit{Ours (Frozen CLIP)}} \\
\textbf{TDS-CLIP-B/16} &CLIP-400M& 8$\times$3$\times$2& 0.2$\times$6 & 60.1 & 85.4 & 71.8 & 93.0\\
\textbf{TDS-CLIP-B/16} &CLIP-400M& 16$\times$3$\times$2& 0.4$\times$6 & 61.0 & 86.3 & 72.1 & 93.3 \\
\textbf{TDS-CLIP-L/14} &CLIP-400M& 8$\times$3$\times$2& 0.8$\times$6 & 63.0 & 87.8 & 73.4 & 93.8\\
\textbf{TDS-CLIP-L/14} &CLIP-400M& 16$\times$3$\times$2& 1.6$\times$6 & \textbf{64.1} & \textbf{88.2} & \textbf{74.1} & \textbf{94.0}\\
\bottomrule
\end{tabular}

\end{table*}

\subsection{Training Loss}

Due to the removal of the text branch, the visual-only task requires that the model can focus more on spatio-temporal modeling of the video action, and thus can be optimized directly using the cross-entropy loss function.
Given an input video, the side network generates a video representation $f_{out} \in \mathbb{R}^{T\times N\times d}$, where $d$ denotes the final linear layer mapping output shape, and then applies Global Average Pooling (GAP) to the patch markers to obtain the final representation $f_{emb}$. This process is denoted as:
\begin{equation}
f_{emb} = GAP(Proj(f_{out})),
\end{equation}
in order to mitigate model overfitting and improve generalization across classes, we introduce label-smooth to regularize constraints on the loss function. The final shape of the loss function combining the label smoothing method is:
\begin{equation}
\mathcal{L}_{\text{CE-LS}} = \left[ (1 - \alpha) \cdot \mathbb{I}(i = y) + \alpha/N \right] \mathcal{L}_{\text{CE}},
\end{equation}
where $\alpha$ is the smoothing intensity, $\mathbb{I}()$ is the exponential function, $y$ is the video category label, and $N$ is the total number of categories used for classification in the dataset. Further, the final loss function representation for training can be obtained by expanding $\mathcal{L}_{\text{CE}}$, where $Y_i$ denotes the category with label smoothing regularization.
\begin{equation}
\mathcal{L}_{\text{CE}} = - \sum_{i=1}^{N} Y_i \log \left( \frac{exp(f_{emb}, i)}{\sum_{j=1}^{N} exp(f_{emb},j)} \right)
\label{eq:cross_entropy}
\end{equation}

\section{Experiments}

\subsection{Experimental Setup}
To evaluate the effectiveness of our method, we assess it on several widely used action recognition benchmark datasets, including Something-Something V1 (SSv1)~\cite{goyal2017something}, Something-Something V2 (SSv2)~\cite{goyal2017something}, and Kinetics400 (K400)~\cite{kay2017kinetics}. SSv1 and SSv2 are commonly used datasets for evaluating the importance of temporal modeling. Kinetics400 is a comprehensive action recognition dataset covering 400 different human actions and is used to evaluate the generalization performance of the methods. In Kinetics400, most action categories are biased toward static scene context. In SSv1 and SSv2, the action categories are less related to the static scene context, but are closely related to the dynamic information in the video.

As in the previous method~\cite{pan2022st,wang2024multimodal}, we use the pre-trained CLIP ViT-B/16 and ViT-L/14 as spatial encoders. By adjusting the dimensions, our model balances memory usage and performance. Due to hardware dilemmas, unless otherwise specified, we use the corresponding ViT-B/16 or ViT-L/14 structure in the temporal encoder to form the TDS-CLIP framework, employing a sparse frame sampling strategy of 8 or 16 frames during both training and inference. For more details on parameter settings and techniques, please see the \emph{Supplementary Material}.

\begin{table*}[h]
\centering
\label{tab:2}

\caption{\centering Comparison of performance with state-of-the-art methods on Kinetics-400. “Cr.” and “Cl.” are the abbreviation for “spatial crops” and “temporal clips”.}
\begin{tabular}{l c c c c c}
\toprule
\textbf{Method}  & \textbf{Pre-train} & \textbf{Frames$\times$Cr.$\times$Cl.}  & \textbf{TFLOPs} &\textbf{Top-1(\%)} & \textbf{Top-5(\%)}\\

\midrule
\multicolumn{6}{l}{\textit{Full Fine-tuning}} \\
Video Swim-L~\cite{liu2022video}& IN-21K &32$\times$3$\times$4&0.6$\times$12&83.1&95.9\\
ViViT FE-L~\cite{arnab2021vivit} & IN-21K &128$\times$3$\times$1&4.0$\times$3&81.7&93.8\\
MViT-B~\cite{fan2021multiscale}& Kinetics-600 &32$\times$5$\times$1&0.2$\times$5&82.9&95.7\\
Uniformer V2-B/16~\cite{li2023uniformerv2}& CLIP-400M &8$\times$3$\times$4&0.2$\times$12&85.6&97.0\\
BIKE-L/14~\cite{wu2023bidirectional} & CLIP-400M &16$\times$3$\times$4&0.8$\times$12&\textbf{88.1}&\textbf{97.9}\\
ActionCLIP-B/16~\cite{wang2023actionclip} & CLIP-400M &32$\times$3$\times$10&0.6$\times$30&83.8&96.2\\
X-CLIP-B/16~\cite{wang2023actionclip} & CLIP-400M &16$\times$3$\times$4&0.6$\times$12&84.7&96.8\\
S-ViT-B/16~\cite{zhao2023streaming} & CLIP-400M &16$\times$3$\times$4&0.7$\times$12&84.7&96.8\\
STAN-B/16~\cite{liu2023revisiting} &CLIP-400M&
16$\times$3$\times$1& 0.7$\times$1 & 84.2 & 96.5\\
\midrule
\multicolumn{6}{l}{\textit{Frozen CLIP}} \\
EVL-B/16~\cite{lin2022frozen} &CLIP-400M& 8$\times$3$\times$1& 0.4$\times$3 & 82.9 & -\\
ST-Adapter-B/16~\cite{pan2022st} &CLIP-400M&8$\times$3$\times$1& 0.2$\times$3 &82.0 & 95.7\\
ST-Adapter-B/16~\cite{pan2022st} &CLIP-400M&32$\times$3$\times$1& 0.6$\times$3 &82.7 & 96.2\\
DUALPATH-B/16~\cite{park2023dual} &CLIP-400M& 32$\times$3$\times$1& 0.7$\times$3 & \textbf{85.4} & \textbf{97.1}\\
AIM-B/16~\cite{yang2023aim} &CLIP-400M& 8$\times$3$\times$1& 0.2$\times$3 & 83.9 & 96.3\\
AIM-B/16~\cite{yang2023aim} &CLIP-400M& 32$\times$3$\times$1& 0.6$\times$3 & 84.7 & 96.7\\
DiST-B/16~\cite{qing2023disentangling} &CLIP-400M& 8$\times$3$\times$1& 0.2$\times$3 & 83.6 & 96.3\\
DiST-B/16~\cite{qing2023disentangling} &CLIP-400M& 32$\times$3$\times$1& 0.7$\times$3 & 85.0 & 97.0\\
M\textsuperscript{2}-CLIP-B/16~\cite{wang2024multimodal} &CLIP-400M& 32$\times$3$\times$4& 0.8$\times$12 & 84.1 & 96.8 \\
Side4Video-B/16~\cite{yao2023side4video} &CLIP-400M& 32$\times$3$\times$4& 0.7$\times$12 & 84.2 & 96.5 \\
\midrule
\multicolumn{6}{l}{\textit{Ours (Frozen CLIP)}} \\

\textbf{TDS-CLIP-B/16} &CLIP-400M& 8$\times$3$\times$4& 0.2$\times$12 & 83.9 & 96.1 \\
\textbf{TDS-CLIP-B/16} &CLIP-400M& 32$\times$3$\times$4& 0.7$\times$12 & 84.5 & 96.7\\
\textbf{TDS-CLIP-L/14} &CLIP-400M& 8$\times$3$\times$4& 0.8$\times$12 & 86.7 & 97.5\\
\textbf{TDS-CLIP-L/14} &CLIP-400M& 16$\times$3$\times$4& 1.6$\times$12 & \textbf{87.2} & \textbf{97.6}\\
\bottomrule

\end{tabular}
\vspace{-0.3cm}
\end{table*}

\subsection{Comparison with State-of-the-art}

\subsubsection{Results on Something-Something V1\&V2} Tab.~\hyperref[tab:1]{\uppercase\expandafter{\romannumeral1}} shows the comparison results with other state-of-the-art methods on SSv1 and SSv2. Using both fully fine-tuning and frozen CLIP parameter methods, our approach achieves competitive results on both SSv1 and SSv2. For instance, in the fully fine-tuning approach, UniFormer-V2-L/14~\cite{li2023uniformerv2} achieves the best classification results with the ViT-L/14 backbone, while our TDS-CLIP demonstrates superior performance with the same backbone (ViT-L/14, 16 frames: 64.1\% vs. 60.5\% on SSv1; 74.1\% vs. 72.1\% on SSv2) while using fewer frames. In the frozen CLIP parameter methods, Dist-L/14~\cite{qing2023disentangling} uses the same side network structure; on the contrast, TDS-CLIP achieves better results with the same frame and sampling settings (ViT-L/16, 74.1\% vs. 72.5\% on SSv2), indicating that the proposed adapter enables effective temporal modelling. Additionally, it can be observed that the ViT-L/14 backbone outperforms the ViT-B/16 backbone under the same settings. These experimental results show that the proposed TD-Adapter and SME-Adapter are effective methods for enhancing temporal modeling and facilitating the transfer of knowledge from images to videos in frozen CLIP models.

\subsubsection{Results on Kinetics-400} To evaluate the generalization performance of our method, we conduct experiments on the K400 dataset. Compared to SSv1 and SSv2, the K400 emphasizes learning appearance features. In addition, compared with methods such as DiST~\cite{qing2023disentangling}, TDS-CLIP has no independent modeling structure for spatial branches or, instead, simply relies on CLIP's own powerful spatial modeling capabilities to provide effective spatial features for temporal branches. Therefore, the improvement in K400 dataset is not as obvious as in SSv1 and SSv2, but it still achieves better performance. As shown in Tab.~\hyperref[tab:2]{\uppercase\expandafter{\romannumeral2}} , our method underperforms BIKE-L/14~\cite{wu2023bidirectional} in terms of appearance feature learning capability when compared to fully fine-tuned approaches. However, with 8-frame inputs, our performance surpasses that of ActionCLIP-B/16~\cite{wang2023actionclip}, MViT-B~\cite{fan2021multiscale}, and Video Swin-L~\cite{liu2022video} with 32-frame inputs, while also maintaining lower computational cost. 

When compared to methods that freeze CLIP, our performance with 32-frame inputs is lower than DiST-B/16~\cite{qing2023disentangling}, AIM-B/16~\cite{yang2023aim}, and DUALPATH-B/16~\cite{park2023dual}, which possess independent spatial feature learning mechanisms. In contrast, TDS-CLIP focuses more on temporal feature learning and modeling. Notably, we achieve a competitive advantage when compared to methods like DiST-B/16~\cite{qing2023disentangling}, AIM-B/16~\cite{yang2023aim}, ST-Adapter-B/16~\cite{pan2022st}, and EVL-B/16~\cite{lin2022frozen} with 8-frame inputs, indicating that TDS-CLIP captures more comprehensive video action information with fewer input frames. Lastly, we also evaluate the performance of TDS-CLIP with an 8-frames and 16-frames ViT-L framework, demonstrating that larger-scale frameworks are effective in improving performance. From the comparison results, we can see that the increase in the number of frames has little gain in performance improvement, which may be the reason for the weak timing correlation of the Kinetics400. Due to memory limitations, we test up to 16 frames of performance in the ViT-L/14, but this is enough to illustrate the advantages. The experimental results demonstrate that TDS-CLIP not only excels in datasets with strong temporal correlations but also performs well in datasets with significant appearance background correlations. With these observations, we can conclude that TDS-CLIP has the dual advantage of spatial modeling and temporal modeling, showcasing robust generalization capabilities.

\subsection{Ablation and Analysis}
In this section, unless otherwise specifed, we use ViT-B/16 as the backbone and the 8 input frames on Something-Something V1. The experiments used 8 A100 graphics cards and run in the pytorch framework.

\subsubsection{Study of component effectiveness} We explore the effectiveness of the proposed adapter and the results are shown in Tab.~\hyperref[tab:3]{\uppercase\expandafter{\romannumeral3}}. The conventional 3D Conv is used to replace the TD-Adapter for the baseline of comparison. We can observe that the classification accuracy is improved by 1.0\% after adding TD-Adapter, which proves that TD-Adapter can effectively improve the temporal modelling capability. Then, we add SME-Adapter, and the accuracy is improved even further, which implies that the SME-Adapter guides the side network to learn the motion features to further enhance the temporal modelling capability of the model, and proves that TD-Adapter and SME-Adapter are complementary to each other in terms of performance improvement.

\begin{table}[ht]

    \centering
    \label{tab:3}
    \caption{Study on the location of TD-Adapter layers.}
    \begin{tabular}{cc}
                \toprule
                Component & Top-1(\%) Acc. \\
                \midrule
                Baseline & 58.1 \\
                + TD-Adapter & 59.1  \\
                + SME-Adapter & \textbf{59.4}  \\
                \bottomrule
    \end{tabular}
\end{table}

\subsubsection{The impact of local frame sample number}

We compare the performance impact of different numbers of local sampling video frames, and the results are shown in Tab. ~\hyperref[tab:4]{\uppercase\expandafter{\romannumeral4}}. 1 indicates that the SME-Adapter is not enabled, and serves as the baseline for comparison. The accuracy improves by 0.5\% when the number is increased to 3 and by 1.4\% when the number is increased to 5. This shows that increasing the number of local sampling frames can effectively improve the recognition performance. We do not test larger quantities because the total number of video frames used for testing is not large and bigger quantities may destroy the sparse sampling efficiency.
\begin{table}[ht]
    \centering
    \label{tab:4}
    \caption{Study on the location of TD-Adapter layers.}
    \begin{tabular}{cc}
                \toprule
                Number & Top-1(\%) Acc. \\
                \midrule
                1 & 58.0  \\
                3 &  58.5 \\
                5 &  \textbf{59.4} \\
                
                \bottomrule
    \end{tabular}
\end{table}
\subsubsection{The impact of fusion rate}

The fusion ratio represents the proportion of the proposed method in feature fusion. The test results, as shown in Tab.~\hyperref[tab:5]{\uppercase\expandafter{\romannumeral5}}, indicate that performance is optimal under equal-ratio fusion, while the performance deteriorates when the output feature ratio of the SME-Adapter is either too high or too low. This shows that the appropriate equal-ratio feature fusion method can well exploit the ability of SME-Adapter to learn and capture motion features.
\begin{table}[ht]
    \centering
    \label{tab:5}
    \caption{Study on the location of TD-Adapter layers.}
    \begin{tabular}{ccc}
                \toprule
                $\alpha$ & $\beta$ & Top-1(\%) Acc.\\
                \midrule
                1.0 & - & 58.1 \\
                - & 1.0 & 47.4 \\
                0.5 & 1.0 & 58.4 \\
                1.0 &  1.0 &\textbf{59.4} \\
                1.0 &  0.5 &58.9 \\
                
                \bottomrule
    \end{tabular}
\end{table}

\subsubsection{Ablations for TD-Adapter}

We explore the impact of different core modules of the TD-Adapter on performance, including temporal convolutional layer enhancement~\cite{wang2024multimodal} and direct feature difference calculation, as shown in Tab.~\hyperref[tab:6]{\uppercase\expandafter{\romannumeral6}}. Using pooling layers achieves a classification accuracy of 59.4\%, while temporal convolutional layer enhancement yields a slightly lower accuracy, decreasing by 0.2\%. Direct feature difference calculation yields the lowest accuracy, as this approach tends to cause the loss of spatial features. It is worth noting that pooling is a parameter-free operation, thus helping to reduce the number of parameters with essentially the same performance.

Furthermore, we test different kernel sizes for the pooling layers, specifically k=3, 5, and 7, as illustrated in Tab.~\hyperref[tab:7]{\uppercase\expandafter{\romannumeral7}}. The classification accuracy decreases as the kernel size increases, indicating that an excessively large kernel size may disrupt the temporal relationships between features, therefore, larger kernel sizes should not be selected.
\begin{table}[ht]
    \centering
    \label{tab:6}
    \caption{Study on the location of TD-Adapter layers.}
    \begin{tabular}{ccc}
                \toprule
                Method & Top-1(\%) Acc. & GFLOPs\\
                \midrule
                Pool &  \textbf{59.4}& - \\
                Conv &  59.3& 1.7 \\
                Direct &  57.3& - \\
                \bottomrule
    \end{tabular}
\end{table}
\begin{table}[ht]
    \centering
    \label{tab:7}
    \caption{Study on the location of TD-Adapter layers.}
    \begin{tabular}{cc}
                \toprule
                Kernel Size & Top-1(\%) Acc. \\
                \midrule
                k=3 & \textbf{59.4} \\
                k=5 & 59.2  \\
                k=7 & 58.8  \\
                \bottomrule
    \end{tabular}
\end{table}
\subsubsection{The impact of dimensionality}

In the transformer architecture, adjusting the dimensionality can effectively control the model's memory consumption and impact its performance. Generally, increasing the dimensionality enhances the model's complexity and memory usage, while also improving the modeling capacity of different layers. We follow the baseline comparison methods to test the impact of various dimensionality settings on performance, as shown in Tab.~\hyperref[tab:8]{\uppercase\expandafter{\romannumeral8}}. The results show that the model achieves the best classification performance at 320 dimensions, while both excessively large or small dimensions constrain performance.
\begin{table}[ht]
    \centering
    \label{tab:8}
    \caption{Study on the location of TD-Adapter layers.}
    \begin{tabular}{cc}
                \toprule
                Embedding & Top-1(\%) Acc. \\
                \midrule
                128 &  57.4 \\
                320 &  \textbf{59.4} \\
                512 &  58.2 \\
                
                \bottomrule
    \end{tabular}
\end{table}

\subsubsection{Study on the location of TD-Adapter layers}

We study the location of the TD-Adapter within the structure, with the results shown in Tab.~\hyperref[tab:9]{\uppercase\expandafter{\romannumeral9}}. We experiment with adding the TD-Adapter to the first half of the network (layers 1-6) and the second half (layers 7-12). Replacement with normal 3D convolution when not using TD-Adapter. The table shows that while using more TD-Adapters generally leads to better classification results, adding them to shallower layers provides greater benefits than those obtained from deeper layers. Additionally, the performance is further improved when the SME-Adapter is also used.
\begin{table}[ht]
    \centering
    \label{tab:9}
    \caption{Study on the location of TD-Adapter layers.}
    \begin{tabular}{cccc}
        \toprule
        Layer 1-6 & Layer 7-12 & SME-Adapter & Top-1 Acc. \\
        \midrule
       \Checkmark &  \XSolidBrush & \XSolidBrush & 58.8\% \\
       \XSolidBrush & \Checkmark & \XSolidBrush & 58.6\% \\
        \Checkmark & \Checkmark & \XSolidBrush & 59.1\% \\
        \Checkmark & \Checkmark & \Checkmark & \textbf{59.4\%}\\
        \bottomrule
    \end{tabular}

\end{table}
\begin{figure}[h]
    \centering
    \includegraphics[width=\linewidth]{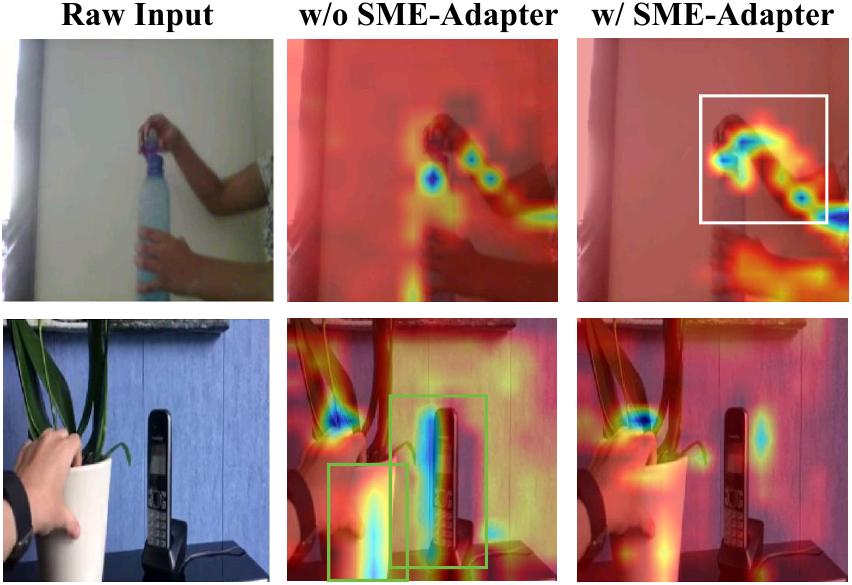}
    \caption{Visualisation of activation maps for image features. Blue colour indicates larger magnitude of feature activation values and red colour indicates smaller magnitude of feature activation values. White and green boxes indicate focus.}
    \label{fig:cam}    
   
\end{figure}

\subsubsection{Comparison of modes}
In this section, we compare the feature models used by several state-of-the-art methods, as shown in Tab.~\hyperref[tab:10]{\uppercase\expandafter{\romannumeral10}}. M\textsuperscript{2}-CLIP~\cite{wang2024multimodal} and DiST~\cite{qing2023disentangling} model both temporal and spatial branches and use text information. In contrast, Side4video~\cite{yao2023side4video} and TDS-CLIP only model temporal branches and do not use text information, which is consistent with the comparison results in Section 4.2 and shows that combining spatial and text information limits certain temporal modeling capabilities but at the same time this is a focus for future research.

\begin{table}[h]
    \centering
    \label{tab:10}
    \caption{Comparison of modes.}
    \begin{tabular}{ccc}
    \toprule
    Method & Mode & Text\\
    \midrule
    M\textsuperscript{2}-CLIP~\cite{wang2024multimodal} &  Temporal+Sptial& \Checkmark  \\
    DiST~\cite{qing2023disentangling} & Temporal+Sptial& \Checkmark  \\
    Side4Video~\cite{yao2023side4video} & Temporal& \XSolidBrush \\
    \midrule
    TDS-CLIP & Temporal& \XSolidBrush \\
    \bottomrule
    \end{tabular}
   
    \label{fig:mode}

\end{table}

\subsubsection{Visualisation Related}
As show in Fig. \ref{fig:cam}, we further visualize the activation maps of features in the side network with and without using the SME-Adapter. It is evident that the SME-Adapter is more sensitive to the motion areas of the target and does not show a bias towards background information (the white box in figure). For example, our SME-Adapter focuses more on the motion region of the hand that interacts with the object, rather than all the targets such as flower pots, water bottles, or mobile phones that appear in the scene (the white green in figure), which is more in line with the human perception. Without the SME-Adapter, the model lacks this motion attention. This indicates that the proposed SME-Adapter facilitates the learning of motion features in videos, enabling the model to focus more accurately on dynamically moving target objects.

\begin{figure}[t]
    \centering
    
    \includegraphics[width=\linewidth]{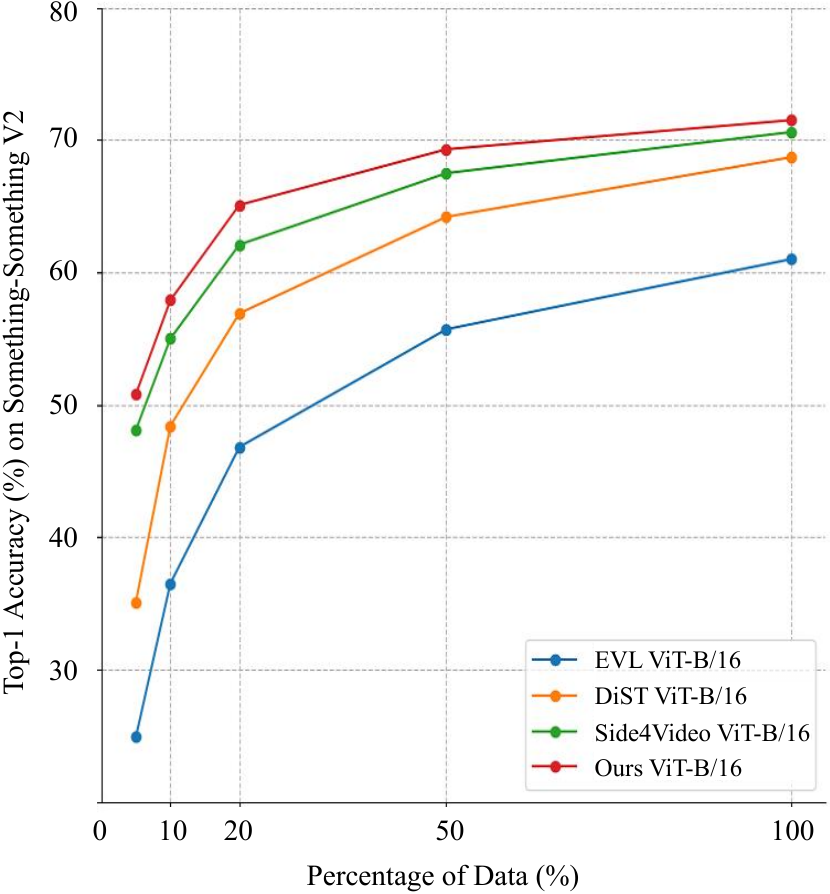}
    \caption{Comparison on data efficiency.}
    \label{fig:data} 
    \vspace{-0.4cm}
\end{figure}

\subsubsection{SME-Adapter Location Experiment}

We test the impact of placing the SME-Adapter at different positions within the network on overall performance. Specifically, we design five modes: (a) Spatial: equipping the SME-Adapter at the input of the CLIP spatial path. (b) Temporal: equipping the SME-Adapter at the input of the side network temporal path. (c) Spatial+Temporal: equipping the SME-Adapter at both the input of the CLIP spatial path and the input of the side network temporal path. (d) Cross Path: connecting the SME-Adapter to the temporal path via a cross-path mode. (e) Additional Path: connecting the SME-Adapter to the temporal path via an additional path mode. The results of these tests are summarized in Tab.~\hyperref[tab:11]{\uppercase\expandafter{\romannumeral11}}. We observe that the best performance is achieved when the SME-Adapter is placed solely at the input of the temporal path, whereas equipping it at the CLIP spatial path results in a significant performance degradation. The Cross Path mode is close to the Temporal mode, and Temporal is the best choice for simplicity reasons.

\begin{table}[ht]
    \centering
    \label{tab:11}
    \caption{Study on the location of TD-Adapter layers.}
    \begin{tabular}{cc}
        \toprule
        Mode &  Top-1(\%) Acc. \\
        \midrule
        Spatial&  53.4 \\
        Temporal&  \textbf{60.1} \\
        Spatial+Temporal&  55.6 \\
        Cross Path&  59.8 \\
        Additional Path&  58.7 \\
        \bottomrule
    \end{tabular}
    
\end{table}
\begin{table}[h]
    \centering
    \label{tab:12}
    \caption{Inference Memory Efficiency. w/ indicates “with".}
    \begin{tabular}{cc}
        \toprule
        Component & Memory(Gib) \\
        \midrule
        Baseline  & 8.334\\
        w/ ST-Adapter & 23.671\\
        w/ 3D Conv & 9.121\\
        \midrule
        w/ TD-Adapter &  8.781\\
        w/ SME-Adapter &  8.799\\
        \bottomrule
    \end{tabular}

\end{table}

\begin{table*}[ht]
\centering
\label{tab:13}
\caption{Configurations for Kinetics-400, Something-Something V1 and Something-Something V2.}
\begin{tabular}{l c c c c c c}
\toprule
dataset   & \multicolumn{2}{c}{Something-Something V1} & \multicolumn{2}{c}{Something-Something V2} &\multicolumn{2}{c}{Kinetics-400}\\

backbone& ViT-B & ViT-L & ViT-B & ViT-L & ViT-B & ViT-L\\
\midrule
dimension size & \multicolumn{6}{c}{320} \\
difference frame  number & \multicolumn{6}{c}{5} \\
pooling kernel shape  & \multicolumn{6}{c}{3 $\times$ 1 $\times$ 1} \\
optimizer & \multicolumn{6}{c}{AdamW, weight decay=0.15, betas=[0.9, 0.999], schedule=Cosine} \\
learning rate & 1e-3 & 1e-4 & 1e-3 & 1e-4 & 1e-3 & 1e-4  \\
batch size &  \multicolumn{4}{c}{128} & \multicolumn{2}{c}{256}\\
epochs & 40 & 30 &\multicolumn{2}{c}{30}&\multicolumn{2}{c}{30}\\
warmup epochs & \multicolumn{6}{c}{4} \\
seed & \multicolumn{6}{c}{1024} \\
\midrule
training resize  & \multicolumn{6}{c}{RandomResizedCrop}\\
label smoothing & \multicolumn{6}{c}{0.1}\\
repeated sampling & \multicolumn{6}{c}{2}\\
random augment & \multicolumn{4}{c}{rand-m7-n4-mstd0.5-inc1}\\
random flip & \multicolumn{6}{c}{0.5}\\
test views & \multicolumn{4}{c}{3 spatial $\times$ 2 temporal}  & \multicolumn{2}{c}{3 spatial $\times$ 4 temporal}\\

\bottomrule
\end{tabular}
\vspace{-0.3cm}
\end{table*}
\subsubsection{Data Efficiency}

Data efficiency refers to fine-tuning a model using only a subset of the training data, thereby maximizing the utility of limited data. Fig.~\ref{fig:data} illustrates the impact of different proportions of the training dataset on the performance of TDS-CLIP ViT-B/16. Our model demonstrates superior data efficiency, as evidenced by the B/16 model outperforming EVL ViT-B/16 \cite{lin2022frozen}, DiST ViT-B/16 \cite{qing2023disentangling}, and Side4Video ViT-B/16 \cite{yao2023side4video} by 25.9\%, 15.7\%, and 2.7\% points, respectively, when using only 5\% of the training data. Furthermore, our model maintains a competitive advantage across other data efficiency scenarios.

\subsubsection{Inference Memory Efficiency}

Tab.~\hyperref[tab:12]{\uppercase\expandafter{\romannumeral12}} summarizes the memory efficiency in the inference phase. The baseline approach uses only the side network structure without any temporal modeling module, while both temporal and spatial branches are used in  ST-Adapter, resulting in a huge memory overhead. From the results, it can be seen that making the proposed method saves a lot of memory.


\subsection{Implementation Details}

Tab.~\hyperref[tab:13]{\uppercase\expandafter{\romannumeral13}} summarises the fine-tuned configurations for several datasets and models, including Something-Something V1 \cite{goyal2017something}, Something-Something V2 \cite{goyal2017something} and Kinetics-400 \cite{kay2017kinetics}. Something-Something is a large-scale video dataset for action recognition, V1 including about 110K videos (V2 contains more) covering 174 fine-grained action categories. Kinetics-400 is a large-scale YouTube video dataset with about 240K training videos and 20K validation videos covering 400 categories, with each clip lasting about 10 seconds.

All experiments in the paper are implemented in PyTorch \cite{paszke2019pytorch}. We use CLIP \cite{clip} with ViT-B/16, ViT-L/14 for video action recognition. Unless otherwise specified, in general, we use simpler data augmentation techniques than end-to-end fine-tuning.

\section{Conclusion}
In this paper, we introduce a novel approach to video action recognition task that balances memory efficiency with effective temporal modeling. Our method addresses the challenging task of transferring large, task-specific vision-language models to the domain of video action recognition, which can be facilitated for researchers with limited hardware. The core innovation of our work is the development of a temporal modeling adapter tailored for side networks and the integration of spatio-temporal features, all while circumventing the need for backpropagation through the CLIP pre-trained model. Comprehensive experiments carried out on various benchmark datasets demonstrate that our method significantly enhances the temporal modeling capabilities. The disadvantage of this approach is that it sacrifices zero-shot generalization capability. In the era of large language models dominated by Transformers structures, we hope our work will inspire further exploration of parameter-efficient fine-tuning methods for video temporal analysis.

\bibliographystyle{IEEEtran}

\vfill

\end{document}